\documentclass[two side,journal]{IEEEtran} 
\usepackage{amsmath,amsfonts} 
\usepackage{algorithmic}
\usepackage{algorithm}  
\usepackage{array} 
\usepackage[caption=false,font=normalsize,labelfont=sf,textfont=sf]{subfig}
\usepackage{textcomp}
\usepackage{stfloats}
\usepackage{url}
\usepackage{verbatim} 
\usepackage{graphicx}
\usepackage{gincltex}
\usepackage{lipsum} 
\usepackage{cite}
\usepackage{amsthm}
\usepackage{pgfplots} 
\pgfplotsset{compat=1.6}
\usepackage{amssymb}
\usepackage{csvsimple}
\usepackage{tikz} 
\usepackage{tikzscale}
\usepackage{tabularx}
\usepackage{booktabs} 
\usetikzlibrary{shapes,arrows,3d,positioning,calc,spy}  
\newtheorem{remark}{Remark} 
 
\newtheorem{prop}{Proposition}  

\newtheorem{lemma}{Lemma}

\newcolumntype{C}{>{\centering\arraybackslash}X} 
\usepackage{xcolor}

\begin{document}
\title{$\ell_1$-Norm Regularized $\ell_1$-Norm Best-Fit Lines}
\author{Xiao Ling, Paul Brooks, ~\IEEEmembership{Member,~IEEE}}

\markboth{Journal of \LaTeX\ Class Files,~Vol.~14, No.~8, August~2021}{Xiao Ling \MakeLowercase{\textit{et al.}}: $\ell_1$-norm regularized $\ell_1$-norm best-fit lines}

\IEEEpubid{}
\IEEEpubidadjcol  

\maketitle

\begin{abstract}
In this work, we propose an optimization framework for estimating a sparse robust one-dimensional subspace. Our objective is to minimize both the representation error and the penalty, in terms of the $\ell_1$-norm criterion. Given that the problem is NP-hard, we introduce a linear relaxation-based approach. Additionally, we present a novel fitting procedure, utilizing simple ratios and sorting techniques. The proposed algorithm demonstrates a worst-case time complexity of
$O(m^2 n\log n)$ and, in certain instances, achieves global optimality for the sparse robust subspace, thereby exhibiting polynomial time efficiency.
Compared to extant methodologies, the proposed algorithm finds the subspace with the lowest discordance, offering a smoother trade-off between sparsity and fit. Its architecture affords scalability, evidenced by a 16-fold improvement in computational speeds for matrices of 2000x2000 over CPU version. Furthermore, this method is distinguished by several advantages, including its independence from initialization and deterministic and replicable procedures. The real-world example demonstrates the effectiveness of algorithm in achieving meaningful sparsity, underscoring its precise and useful application across various domains.
\end{abstract}

\begin{IEEEkeywords}
$\ell_1$-norm, robust principal component analysis, sparse principal component analysis, outlier insensitive, principal component analysis, linear programming, optimization.
\end{IEEEkeywords}

\section{Introduction}
Principal Component Analysis (PCA) is a prominent method that applies the concept of the best-fit linear subspace to given data, showcasing its widespread utility in data analysis. It finds the orthogonal linear combinations of the data variables that capture the maximal variance. The first principal component (PC) captures the maximum variance, while successive PCs capture the residual variance on the orthogonal components. These components can be computed using singular value decomposition (SVD). Despite its widespread use, PCA faces three main challenges: it is sensitive to outliers, has scalability issues, and sometimes lacks interpretability. PCA's reliance on the sum of squared errors makes it particularly susceptible to outliers, which can exaggerate errors and obscure the underlying pattern of the data. Additionally, the need for batch processing with SVD can limit its scalability, especially with large datasets. The dense principal components also make it difficult to interpret results in high-dimensional spaces. In recent years, there has been a growing interest in developing methods that aim to find a robust and sparse best-fit subspace. A key criterion that has proven to be particularly useful is the $\ell_1$-norm, which, through various formulations, can introduce both sparsity and robustness into the fitting procedures. Specifically, the use of an $\ell_1$-norm penalty on the bases to induce sparsity, along with the adoption of $\ell_1$-norm error or dispersion for subspace fitting to enhance robustness, have been developed. These approaches offer increased resilience against outliers and improved scalability and interpretability.

In addressing the sensitivity to outliers, \cite{ke2005robust} defined the robust subspace estimation as an $\ell_1$-norm minimization problem, employing alternating convex programming methods. However, the approach is limited by its propensity to converge to local and its dependence on the choice of initial values. \cite{hubert2005robpca} explored robust principal components through a heuristic approach that combines the minimum covariance determinant procedure with projection pursuit, albeit at increased computational costs. \cite{Kwak2008PrincipalMaximization} developed a PCA variant that aims to maximize the $\ell_1$-norm of the spread of points in one direction using a greedy algorithm, but it is prone to local optima. The study by \cite{6126034} introduced the concept of outlier pursuit, solving a similar problem to identify true subspaces in corrupted data, employing proximal gradient algorithms despite sensitivity to problem conditioning and the need for parameter tuning. The seminal study by \cite{Candes2011RobustAnalysis} uncovered robust subspaces using the augmented Lagrange multiplier (ALM) algorithm for a relaxed low-rank approximation problem, yet highlighted its nonscalability.  \cite{markopoulos2014optimal} introduced and proved a polynomial procedure for solving the $\ell_1$ dispersion maximization problem, which also faced limitations in scalability. \cite{Song2017LowError,chier2017} introduced the initial approximation algorithms that come with verifiable guarantees for entrywise $\ell_1$-norm low rank approximation. Their method operates within polynomial time. \cite{Tsagkarakis2017OnRank} proposed a polynomial-time heuristic scheme for approximating a subspace with low representation error in the $\ell_1$-norm. 
Building upon the foundational work of \cite{brooks2013pure}, which introduced a polynomial-time heuristic algorithm for minimizing relaxed $\ell_1$-norm representation errors, the approach presents the first scalable algorithm in this domain.

In the context of sparsity, \cite{jolliffe2003modified} obtained sparse components by constraining the $\ell_1$-norm of PCA components less than a predetermined parameter. \cite{d2004direct} proposed approximating the sparse best-fit line by solving a computationally intensive semidefinite programming relaxation of a variance maximization problem. \cite{zou2006sparse} formulated a sparse PCA problem as a naive elastic net regression on PCA components, searching for sparse linear combinations.  \cite{shen2008sparse} promoted sparsity in loading vectors by adding an $\ell_1$-norm penalty and solved the problem using an iterative method. These approach encountered difficulties in terms of the high computational cost and lack of clear guidance to choose the parameter.

Adding both robustness and sparsity simultaneously is more challenging. \cite{meng2012improve} sought robust sparse PC by maximizing the $\ell_1$-norm dispersion and enforcing a threshold constraint concurrently.\cite{croux2013robust} employed a grid search algorithm to sequentially approximate robust sparse PC, maximizing the dispersion along with the $\ell_1$-norm penalty. \cite{hubert2016sparse} applied \cite{jolliffe2003modified}'s methods on the robust PC derived in \cite{hubert2005robpca}.

In this work, we propose an algorithm with modest computational requirements for $\ell_1$ regularization with the traditional squared $\ell_2$-norm error replaced by the $\ell_1$-norm loss. Consider the optimization problem to find an $\ell_1$-norm regularized $\ell_1$-norm best-fit one-dimensional subspace: 
\begin{align}
\label{formulation1}
    &\min_{v,\alpha} \sum_{i\in N}\|x_i - v\alpha_i\|_1 + \lambda\|v\|_1,
\end{align}
where $x_i$, $i \in N$ are points in $\mathbb{R}^m$. An optimal vector $v^*$ determines a line through the origin corresponding to the best-fit subspace. For each point $x_i$, the optimal coefficient $\alpha_i^*$ specifies the locations of the projected points $v^*\alpha_i$ on the line defined by $v^*$. Due to the nature of the $\ell_1$ norm, some components of $v$ will be reduced to zero if $\lambda$ is large enough. Therefore, our proposed method simultaneously generates both a best-fit and a sparse line in $m$ dimensions, which makes it suitable for large or high-dimensional data. The method can be extended to the problem of fitting subspaces. The problem in \eqref{formulation1} is non-linear, non-convex, and non-differentiable. Therefore, we adapt the approximation algorithm of \cite{Brooks2017IdentifyingRanking} to the regularized problem. Simulations and applications to real data demonstrate that the method is efficient for higher-dimensional data and capable of finding the best possible sparse bases of the robust subspace.

\section{Estimating an $\ell_1$-Norm Regularized $\ell_1$-Norm Best-Fit Line}
In this section, we will extend the sorting method introduced in \cite{Brooks2017IdentifyingRanking} for estimating $\ell^1$-norm best-fit lines to the setting where we add a penalty for sparsity. \cite{gillis2018} demonstrated that finding an $\ell_1$-norm best-fit line is NP-hard.  First, we introduce four sets of goal variables $\epsilon_{ij}^+,\epsilon_{ij}^-$ and $\zeta_j^+,\zeta_j^-$.  The optimization problem in $\eqref{formulation1}$ can be recast as the following constrained mathematical program.
    \begin{align}
        \label{formulation2}
         \displaystyle\min_{\genfrac{}{}{0pt}{2}{v\in \mathbb{R}^m,\alpha \in \mathbb{R}^n}{ \genfrac{}{}{0pt}{2}{\epsilon^+, \epsilon^- \in \mathbb{R}^{n\times m}_+,}{\zeta^+, \zeta^- \in \mathbb{R}^m_+}} } &   \sum_{i\in N}\sum_{j\in M} (\epsilon_{ij}^+ + \epsilon_{ij}^-) + \lambda\sum_{j \in M} (\zeta_j^+ + \zeta_j^-), 
    \end{align}
s.t.
    \begin{align*} 
         v_j\alpha_i+\epsilon_{ij}^+-\epsilon_{ij}^-&=x_{ij}, i\in N, j\in M, \\
         v_j + \zeta_j^+ - \zeta_j^-&=0, j \in M,\\
         \epsilon_{ij}^+, \epsilon_{ij}^-, \zeta_j^+, \zeta_j^-  & \geq 0, i \in N, j\in M. 
    \end{align*} 
\begin{prop}
The formulation \eqref{formulation2} is equivalent to \eqref{formulation1}.
\end{prop}  
\begin{proof}
The presence of absolute values in the objective function can be avoided by replacing each $x_{ij}-v_j\alpha_i$ with $\epsilon_{ij}^+-\epsilon_{ij}^-, \epsilon_{ij}^+,\epsilon_{ij}^-\geq 0$ and each $v_j$ with $\zeta_j^+-\zeta_j^-,\zeta_j^+,\zeta_j^-\geq 0$, and these become the constraints. The new objective function $\sum_{i=1}^n\sum_{j=1}^m|\epsilon_{ij}^+ - \epsilon_{ij}^-| + \lambda |\zeta_j^+ - \zeta_j^-|$ can be replaced with $\sum_{i=1}^n\sum_{j=1}^m(\epsilon_{ij}^+ + \epsilon_{ij}^-) + \lambda (\zeta_j^+ + \zeta_j^-)$. This problem has an optimal solution with at least one of the values in each of the pairs $\epsilon_{ij}^+, \epsilon_{ij}^-$ and $\zeta^+, \zeta^-$ at zero. In that case, $x_{ij}-v_j\alpha_i=\epsilon_{ij}^+$, if $x_{ij}-v_j\alpha_i>0$, and $x_{ij}-v_j\alpha_i=-\epsilon_{ij}^-$, if $x_{ij}-v_j\alpha_i<0$. $v_j=\zeta_j^+$, if $v_j>0$, and $v_j=-\zeta_j^-$, if $v_j<0$. Any feasible solution for \eqref{formulation2} generates an objective function value which is the same as that of \eqref{formulation1} using the same values for $v$ and $\alpha$, and vice-versa. Therefore, an optimal solution to \eqref{formulation1} generates a feasible solution for \eqref{formulation2} and vice-versa.
\end{proof} 

An optimal solution to \eqref{formulation2} will be a vector $v^* \in \mathbb{R}^m$, along with scalars $\alpha_i^*$, $i\in N$.  For each point $i$ and feature $j$, the pair ($\epsilon_{ij}^{+*}$,$\epsilon_{ij}^{-*}$) reflects the distance along the unit direction $j$ between the point and its projection.  The pairs $(\zeta_j^{+*}, \zeta_j^{-*})$ provide the difference from zero for each coordinate of $v^*$.

The following proposition provides a foundation for the sorting method proposed by \cite{Brooks2017IdentifyingRanking}.

\begin{prop}
\label{p2}
{\bf \cite{Brooks2019ApproximatingL1}} Let $v\neq$ 0 be a given vector in $\mathbb{R}^m$. Then there is an $\ell_1$-norm projection from the point $x_i\in \mathbb{R}^m$ on the line defined by v that can be reached using at most $m-1$ unit directions.  Moreover, if the preserved coordinate is $\hat{\jmath}$ and $x_{i\hat{\jmath}} \neq 0$, then $v_{\hat{\jmath}} \neq 0$.
\end{prop}
\begin{proof}
A proof is in \cite{Brooks2019ApproximatingL1}.
\end{proof}
 
The idea is to impose the preservation of the same coordinate, $\hat{\jmath}$, in the projections of all points. \cite{Tsagkarakis2017OnRank,chier2017} also propose methods for subspace estimation based on the assumption that all points preserve the same unit directions.  In this work, we are focused on line fitting and we are adding a regularization term to promote sparsity.  

For a line, preservation of the same coordinate means that each point will use the same $m-1$ unit directions to project onto the line defined by $v$. 

By Proposition \ref{p2}, if $x_{\hat{\jmath}}\neq 0$, then $v_{\hat{\jmath}} \neq0$. Therefore, we can set $v_{\hat{\jmath}}=1$ and set $\alpha_i=x_{i\hat{\jmath}}$ to preserve $\hat{\jmath}$ without loss of generality for the error term, though the regularization term is affected.  

\begin{remark}
For any solution to \eqref{formulation1}, a solution with a better objective function can be obtained by dividing $v$ by a large constant.  Therefore, $v$ needs to be normalized in magnitude in some way.  We will do so by requiring $v_{\hat{\jmath}} = 1$.   
\end{remark}

The remaining components of $v$ can be found by solving an LP based on \eqref{formulation2} after replacing $\alpha_i$ with $x_{i\hat{\jmath}}$:
\begin{align}
        \label{formulation3}
        z_{\hat{\jmath}}(\lambda) =\displaystyle\min_{\genfrac{}{}{0pt}{2}{v\in \mathbb{R}^m,v_{\hat{\jmath}}=1 }{\genfrac{}{}{0pt}{2}{\epsilon^+, \epsilon^- \in \mathbb{R}^{n\times m},}{\zeta^+, \zeta^- \in \mathbb{R}^m}}}  & \sum_{i\in N}\sum_{j\in M} (\epsilon_{ij}^+ + \epsilon_{ij}^-) + \lambda \sum_{j\in M}(\zeta_j^+ + \zeta_j^-), 
    \end{align}
s.t.
    \begin{align*} 
        v_jx_{i\hat{\jmath}}+\epsilon_{ij}^+-\epsilon_{ij}^-&=x_{ij}, i\in N, j\in M;j\neq \hat{\jmath}, \\
        v_j + \zeta_j^+ - \zeta_j^-&=0, j\in M,\\
         \epsilon_{ij}^+, \epsilon_{ij}^-, \zeta_j^+, \zeta_j^-  & \geq 0, i \in N, j\in M.
    \end{align*}
Each of the $n$ data points generates $m-1$ constraints in this LP.  

Allowing each coordinate $j$ to serve as the preserved coordinate $\hat{\jmath}$ produces $m$ LPs.  By solving these $m$ LPs and selecting the vector $v$ from the solutions associated with the smallest value of the objective function, we will have the $\ell_1$-norm regularized $\ell_1$-norm best-fit line under the assumption that all points project by preserving the same coordinate $\hat{\jmath}$ and $v_{\hat{\jmath}}=1$.  The following lemma describes how to generate solutions to the LPs by sorting several ratios.
\begin{lemma}
    \label{p4}
    For data $x_i \in \mathbb{R}^m$, $i\in N$, and for a $\lambda \in \mathbb{R}$, an optimal solution to \eqref{formulation3} can be constructed as follows.  If $x_{i\hat{\jmath}} = 0$ for all $i$, then set $v = 0$.  Otherwise, set $v_{\hat\jmath}=1$ and for each $j \neq \hat{\jmath}$,
    \begin{itemize}
        \item Take points $x_i$, $i\in N$ such that $x_{i\hat{\jmath}} \neq 0$ and sort the ratios $\displaystyle\frac{x_{ij}}{x_{i\hat{\jmath}}}$ in increasing order. 
        \item If there is a point $\tilde{\imath}$ where 
       \begin{equation} 
		    \label{solncond}
		    \left|\mbox{sgn}\left(\frac{x_{\tilde{\imath}j}}{x_{\tilde{\imath}\hat{\jmath}}}\right) \lambda +  \sum_{\genfrac{}{}{0pt}{2}{i \in N:}{i < \tilde{\imath}}} |x_{i\hat{\jmath}}| - \sum_{\genfrac{}{}{0pt}{2}{i \in N:}{i > \tilde{\imath}}} |x_{i\hat{\jmath}}| \right| \leq  |x_{\tilde{\imath}\hat{\jmath}}|,
		  \end{equation} 
    then set $v_j = \displaystyle\frac{x_{\tilde{\imath}j}}{x_{\tilde{\imath}\hat{\jmath}}}$.
               \item If no such $\tilde{\imath}$ exists, then set $v_j = 0$.
    \end{itemize}
\end{lemma}
\begin{proof} 
    The problem \eqref{formulation3} is separable into $m$ independent small sub-problems, one for each column $j$. 
    For a given $j$, there is an LP of the form
    \begin{align}
        \label{l1reglp}
        \min_{\displaystyle\genfrac{}{}{0pt}{2}{v_j, \epsilon^+, \epsilon^-, \lambda}{ \zeta^+, \zeta^-}}  & \sum_{i\in N} (\epsilon_{ij}^+ + \epsilon_{ij}^-) + \lambda (\zeta_j^+ + \zeta_j^-), \\
        \mbox{s.t. } &         v_j x_{i\hat{\jmath}}+\epsilon_{ij}^+-\epsilon_{ij}^- =  x_{ij}, i\in N, \nonumber \\
        & v_j + \zeta_j^+ - \zeta_j^- = 0, \nonumber \\
        & \epsilon_{ij}^+, \epsilon_{ij}^-,\zeta_j^{+}, \zeta_j^- \geq 0, i \in N. \nonumber
    \end{align}
    
    
    We will show that the solution for $v_j$ stated in Lemma \ref{formulation1} is optimal by constructing a dual feasible solution that is complementary to the proposed primal feasible solution.  Suppose that the ratios $\frac{x_{ij}}{x_{i\hat{\jmath}}}$, $i\in N$, are sorted in increasing order.  
    
    The dual linear program to \eqref{l1reglp} is
    \begin{align}
        \max_{\pi, \gamma} & \sum_{i \in N} \frac{x_{ij}}{x_{i\hat{\jmath}}} \pi_{i}, \\
        \mbox{s.t. } & \sum_{i \in N} \pi_{i} + \gamma = 0, \\
        & -|x_{i\hat{\jmath}}| \leq \pi_{i} \leq |x_{i\hat{\jmath}}|, i \in N, \label{pibound} \\
        & -\lambda \leq \gamma \leq \lambda.
    \end{align}
    Suppose there is an $\tilde{\imath}$ satisfying \eqref{solncond}.  Then let $\gamma = -\mbox{sgn} \left(\frac{x_{\tilde{\imath}j}}{x_{\tilde{\imath}{\hat{\jmath}}}}\right)\lambda$ and let
    \[
        \pi_i = \left\{\begin{array}{rl} |x_{i\hat{\jmath}}| & \mbox{if $i > \tilde{\imath}$},\\
                                - |x_{i\hat{\jmath}}| & \mbox{if $i < \tilde{\imath}$},\\
                                - \gamma - \sum_{i \neq \tilde{\imath}} \pi_i & \mbox{if $i = \tilde{\imath}$}.
        \end{array}\right.
    \]
    This solution satisfies complementary slackness.  To show that the solution is dual feasible, we need to show that $\pi_{\tilde{\imath}}$ satisfies the bounds in \eqref{pibound} (all other bounds and constraints are satisfied):  
        \begin{eqnarray}
            |\pi_{\tilde{\imath}}| & = & |-\gamma - \sum_{i \neq \tilde{\imath}} \pi_i|, \\
                                 & = & \left|\mbox{sgn} \left(\frac{x_{\tilde{\imath}j}}{x_{\tilde{\imath}{\hat{\jmath}}}}\right)\lambda  + \sum_{i: i < \tilde{\imath}} |x_{i\hat{\jmath}}| - \sum_{i: i > \tilde{\imath}} |x_{i\hat{\jmath}}| \right|,   \\
                                 & \leq & |x_{\tilde{\imath}\hat{\jmath}}|.
        \end{eqnarray}
  The inequality is due to \eqref{solncond}.  There is a complementary dual feasible solution, so the proposed solution must be optimal.  
 %
 
Now suppose that there is no $\tilde{\imath}$ satisfying \eqref{solncond}.  
Note that if \eqref{solncond} is satisfied for some $\tilde{\imath}$ with $\mbox{sgn}\left(\frac{x_{\tilde{\imath}j}}{x_{\tilde{\imath}\hat{\jmath}}}\right) = +$, then
\begin{align}
  \lambda & \geq \sum_{i: i > \tilde{\imath}} |x_{i\hat{\jmath}}| - \sum_{i: i \leq \tilde{\imath}} |x_{i \hat{\jmath}}|, \label{lambdalow}\\
  \lambda & \leq \sum_{i: i \geq \tilde{\imath}} |x_{i\hat{\jmath}}| - \sum_{i: i < \tilde{\imath}} |x_{i \hat{\jmath}}|. \label{lambdahigh}
\end{align}
If \eqref{solncond} is violated for each $\tilde{\imath}$, then for each possible $\tilde{\imath}$ either the lower bound \eqref{lambdalow} or the upper bound \eqref{lambdahigh} for $\lambda$ is violated.  If for a given $\tilde{\imath}$, the lower bound \eqref{lambdalow} is violated, then $\lambda  <  \sum_{i: i > \tilde{\imath}} |x_{i\hat{\jmath}}| - \sum_{i: i \leq \tilde{\imath}} |x_{i \hat{\jmath}}|$.  This implies that the upper bound \eqref{lambdahigh} is satisfied.  If we now consider point $\tilde{\imath} + 1$, then the upper bound is the same as the lower bound for $\tilde{\imath}$ and is therefore satisfied.  So $\lambda$ must violate the lower bound for $\tilde{\imath} + 1$, and we can consider $\tilde{\imath} + 2$ and so on.  Then lower bound is violated for all points with $\mbox{sgn}\left(\frac{x_{\tilde{\imath}j}}{x_{\tilde{\imath}\hat{\jmath}}}\right) = +$, in particular the largest, and so $\lambda < 0$, contradicting the choice of $\lambda$.  A symmetric argument  holds for $\tilde{\imath}$ with $\mbox{sgn}\left(\frac{x_{\tilde{\imath}j}}{x_{\tilde{\imath}\hat{\jmath}}}\right) = -$.  
Therefore, $\lambda > \sum_{i: i \geq \tilde{\imath}}|x_{i\hat{\jmath}}| - \sum_{i: i < \tilde{\imath}} |x_{i\hat{\jmath}}|$, for every $\tilde{\imath}$ with $\mbox{sgn}\left(\frac{x_{\tilde{\imath}j}}{x_{\tilde{\imath}\hat{\jmath}}}\right) = +$ and $\lambda > \sum_{i: i \leq \tilde{\imath}}|x_{i\hat{\jmath}}| - \sum_{i: i > \tilde{\imath}} |x_{i\hat{\jmath}}|$ for every $\tilde{\imath}$ with $\mbox{sgn}\left(\frac{x_{\tilde{\imath}j}}{x_{\tilde{\imath}\hat{\jmath}}}\right) = -$.  In particular, 
\begin{equation}
  \lambda > 
  \left|\sum_{i: \frac{x_{ij}}{x_{i\hat{\jmath}}} < 0} |x_{i\hat{\jmath}}|
  - \sum_{i: \frac{x_{ij}}{x_{i\hat{\jmath}}} > 0} |x_{i\hat{\jmath}}| \right|.
\end{equation}
A dual feasible and complementary solution is to set 
    \[
      \pi_i = \left\{\begin{array}{rl} |x_{i\hat{\jmath}}| & \mbox{if} \frac{x_{ij}}{x_{i\hat{\jmath}}} > 0,\\
	- |x_{i\hat{\jmath}}| & \mbox{if} \frac{x_{ij}}{x_{i\hat{\jmath}}} < 0,
        \end{array}\right.
    \]
and $\gamma = \left|\sum_{i: \frac{x_{ij}}{x_{i\hat{\jmath}}} < 0} |x_{i\hat{\jmath}}|
  - \sum_{i: \frac{x_{ij}}{x_{i\hat{\jmath}}} > 0} |x_{i\hat{\jmath}}| \right|$.  Note that $|\gamma| < \lambda$ by the development above, so the solution is dual feasible and therefore optimal.  
 \end{proof}
 
Given a penalty $\lambda$, an optimal solution to \eqref{formulation3} with the preservation of one coordinate $\hat{\jmath}$ and $v_{\hat{\jmath}} = 1$ requires the sorting of $(m-1)$ lists of ratios according to Lemma \ref{p4}. The process is repeated for each choice of $\hat{\jmath}$ and the solution with the smallest objective function value is retained. Therefore, for a penalty $\lambda$, the proposed method requires sorting $m(m-1)$ lists of ratios in total, each costing $(n\log n)$ running time.  This motivates an $O{(m^2n \log n)}$ algorithm for estimating $v$ formalized in Algorithm 1. 
\begin{prop}
For a given $\lambda$ and data $x_i \in \mathbb{R}^m$, $i\in N$, Algorithm 1 finds an optimal solution to \eqref{formulation3}. 
\end{prop}   
\begin{proof}
For each fixed coordinate, Algorithm 1 finds an optimal solution according to Lemma 1.  From among those solutions, Algorithm 1 picks the one with the smallest combination of error plus regularization term.
\end{proof}

\begin{algorithm}[!htbp]
 \caption{Estimating an $\ell_1$-norm regularized $\ell_1$-norm best-fit line $v^*$ for given $\lambda$.}
 \algsetup{indent=2em} 
 \begin{algorithmic}[1]
 \renewcommand{\algorithmicrequire}{\textbf{Input:}}
 \renewcommand{\algorithmicensure}{\textbf{Output:}} 
 \newcommand{\algorithmicbreak}{\textbf{break}}
 \newcommand{\BREAK}{\STATE \algorithmicbreak}
 \REQUIRE $x_i \in \mathbb{R}^{m}$ for $i = 1,\dotso,n$. $\lambda$.
 \ENSURE $v^*$
  \STATE{Set $z^*=\infty$}
   \FOR {$\hat{\jmath}\in M$}
    \STATE{Set $v_{\hat{\jmath}} = 1$.} 
  \FOR{$j\in M: j \neq\hat{\jmath}$}
  \STATE {Set $v_j = 0$.}
  \STATE{Sort $\left\{\frac{x_{ij}}{x_{i\hat{\jmath}}}: i\in N, x_{i\hat{\jmath}} \neq 0\right\}$.}  
    \FOR{$\tilde{\imath} \in N: x_{\tilde{\imath}\hat{\jmath}} \neq 0$} 
      \IF{$\mbox{sgn}\left(\frac{x_{ij}}{x_{i\hat{\jmath}}}\right)\lambda\in(\sum\limits_{i: i > \tilde{\imath}} |x_{i\hat{\jmath}}|-\sum\limits_{i: i \leq \tilde{\imath}} |x_{i\hat{\jmath}}|,\sum\limits_{i: i \geq \tilde{\imath}} |x_{i\hat{\jmath}}|-\sum\limits_{i: i < \tilde{\imath}} |x_{i\hat{\jmath}}|]$}  
          \STATE Set $v_j=\frac{x_{\tilde{\imath}j}}{x_{\tilde{\imath}\hat{\jmath}}}$. 
       \ENDIF
      \ENDFOR 
  \ENDFOR
  \STATE {set $z=\sum\limits_{i\in N}\sum\limits_{j\in M}|x_{ij}-v_j x_{i\hat{\jmath}}|+\lambda\sum\limits_{j\in M}|v_{j}|$}
  \IF{$z<z^*$}
  \STATE{Set $z^* = z$, $v^{*}=v$}
  \ENDIF
  \ENDFOR
  \RETURN {$v^*$}
 \end{algorithmic} 
 \label{alg1}
 \end{algorithm} 
 
Algorithm 1 finds the best solution that preserves each coordinate $\hat{\jmath}$ and $v_{\hat{\jmath}}$=1 for a given value of $\lambda$. Algorithm 2 seeks the intervals constructed by successive breakpoints - values for $\lambda$ at which the solution is going to change and the conditions of Lemma 1 are satisfied. Algorithm 2 does not determine which coordinate $\hat{\jmath}$ is best to preserve for each interval.
 Algorithm 3 iterates through each interval for $\lambda$ from Algorithm 2 and finds the intervals where preserving $\hat{\jmath}$ minimizes the objective function value.
 

\begin{algorithm} [!htbp] 
 \caption{Find all major breakpoints.}
 \begin{algorithmic}[1]
 \renewcommand{\algorithmicrequire}{\textbf{Input:}}
 \renewcommand{\algorithmicensure}{\textbf{Output:}}
 \REQUIRE $x_i \in \mathbb{R}^{m}$ for $i = 1,\dotso,n$.
 \ENSURE Ordered breakpoints for the penalty $\Lambda$ and solutions $v^{\hat{\jmath}}(\lambda)$ for each choice of preserved coordinate $\hat{\jmath}$, and each $\lambda \in \Lambda$.
 \STATE Set $\Lambda = \{0, \infty\}$.
  \FOR {$\hat{\jmath}\in M$}
    \STATE{Set $v^{\hat{\jmath}}_{\hat{\jmath}} = 1$.}
  \FOR{$j\in M: j \neq\hat{\jmath}$}
  \STATE{Set $\lambda^{\max} = 0$.}
  \STATE{Sort $\left\{\frac{x_{ij}}{x_{i\hat{\jmath}}}: i\in N, x_{i\hat{\jmath}} \neq 0\right\}$.}
    \FOR{$\tilde{\imath} \in N: x_{\tilde{\imath}\hat{\jmath}} \neq 0$}
       \STATE Set {$\lambda = \mbox{sgn}\left(\frac{x_{\tilde{\imath}j}}{x_{\tilde{\imath}\hat{\jmath}}}\right) \left(\sum\limits_{i: i > \tilde{\imath}} |x_{i\hat{\jmath}}|-\sum\limits_{i: i < \tilde{\imath}} |x_{i\hat{\jmath}}|\right) - |x_{\tilde{\imath}\hat{\jmath}}|$}
       \IF{$\lambda + 2|x_{\tilde{\imath}\hat{\jmath}}| > 0$,}  
         \STATE Set $\Lambda = \Lambda \cup \max\{0,\lambda\}$.
         \STATE Set $v^{\hat{\jmath}}_j(\max\{0,\lambda\}) = 
         \frac{x_{\tilde{\imath}j}}{x_{\tilde{\imath}\hat{\jmath}}}$.
       \ENDIF
       \IF{$\lambda +2|x_{\tilde{\imath}\hat{\jmath}}| > \lambda^{\max}$,}
         \STATE{Set $\lambda^{\max} = \lambda + 2|x_{\tilde{\imath}\hat{\jmath}}|$.}
       \ENDIF
     \ENDFOR
     \STATE{Set $\Lambda = \Lambda \cup \{\lambda^{\max}\}$.}
     \STATE{Set $v_j^{\hat{\jmath}}(\lambda^{\max}) = 0$.}
  \ENDFOR
  \ENDFOR 
  \STATE Sort $\Lambda$.
  \RETURN $\Lambda$, $\{v_j^{\hat{\jmath}}(\lambda): j \in M, \hat{\jmath} \in M, \lambda \in \Lambda\}$
 \end{algorithmic} 
 \label{alg2}
 \end{algorithm}  
 
 It is necessary to ``merge'' the intervals for each possible preserved coordinate $\hat{\jmath}$ and determine when the preservation of each coordinate results in the lowest value of the objective function.  Therefore, we need Algorithm 3 to check each consecutive interval for $\lambda$ from Algorithm 2 to determine if changing the preserved coordinate $\hat{\jmath}$ can reduce the objective function value, which may result in new breakpoints that were not discovered using Algorithm 2. 
\begin{algorithm} [!htbp] 
 \caption{Solution Path for $\ell_1$-norm Regularized $\ell_1$-norm best-fit line}
 \begin{algorithmic}[1]
 \renewcommand{\algorithmicrequire}{\textbf{Input:}}
 \renewcommand{\algorithmicensure}{\textbf{Output:}} 
 \REQUIRE A ordered set of breakpoints for the penalty  $(\lambda^k: k=1,\ldots, K)$ and solutions $v^{\hat{\jmath}}(\lambda^k)$ for each choice of preserved coordinate $\hat{\jmath}$. 
 \ENSURE Breakpoints for the penalty $\Lambda$ and solutions $v^*(\lambda)$ for each $\lambda \in \Lambda$. 
 \STATE{Set $\Lambda = \emptyset.$}
  \FOR{$k = 1, \ldots, K-1$}
    \FOR{$\hat{\jmath} \in M$}
      \STATE{Set $z^{\hat{\jmath}}(\lambda^k) = \sum_{i \in N} \|x_i - v^{\hat{\jmath}}x_{i\hat{\jmath}}\|_1 + \lambda^k \|v^{\hat{\jmath}}(\lambda^k)\|_1$}
    \ENDFOR
    \FOR{$j \in M$} 
      \STATE $\beta_L = 
      \begin{aligned}[t]
         &\left\{ \max \frac{z^{j}(\lambda^k) - z^{\hat{\jmath}}(\lambda^k)}{\|v^{\hat{\jmath}}(\lambda^k)\|_1 - \|v^j(\lambda^k)\|_1}:\right. \\
         &\left. \hat{\jmath} \vphantom{\bigg\{} \in M, \|v^j(\lambda^k)\|_1 < \|v^{\hat{\jmath}}(\lambda^k)\|_1  \right\}
      \end{aligned}$
      \STATE $\beta_U = 
      \begin{aligned}[t]
        &\left\{ \{\min \frac{z^{\hat{\jmath}}(\lambda^k) - 
        z^{j}(\lambda^k)}{\|v^{j}(\lambda^k)\|_1 - 
        \|v^{\hat{\jmath}}(\lambda^k)\|_1}:\right.  \\
        &\left. \hat{\jmath} \vphantom{\bigg\{} \in M, \|v^j(\lambda^k)\|_1 > \|v^{\hat{\jmath}}(\lambda^k)\|_1 \right\} 
      \end{aligned}$
      \IF{$\left|\left\{\hat{\jmath}: z^j(\lambda^k) > z^{\hat{\jmath}}(\lambda^k), \|v^j(\lambda^k)\|_1 = \|v^{\hat{\jmath}}(\lambda^k)\|_1\right\}\right| = 0,$}
        \IF{$0< \beta_L < \beta_U$ and $\lambda^k + \beta_L \leq \lambda^{k+1}$, }
        \STATE{Set $\Lambda = \Lambda \cup \{\lambda^k + \beta_L\}$}
        \STATE{Set $v^*(\lambda^k + \beta_L) = v^j(\lambda^k)$}
        \ELSIF{$\beta_L \leq 0 < \beta_U$,}
          \STATE{Set $\Lambda = \Lambda \cup \{\lambda^k\}$}
          \STATE{Set $v^*(\lambda^k) = v^j(\lambda^k)$}
        \ENDIF
      \ENDIF
    \ENDFOR
  \ENDFOR
  \RETURN $\Lambda$, $\{v^*(\lambda): \lambda \in \Lambda\}$
 \end{algorithmic} 
 \label{alg3}
 \end{algorithm}
 
\begin{prop}
\label{p5} 
For data $x_i \in \mathbb{R}^m$, $i\in N$, Algorithms 2 and 3 generate the entire solution path for \eqref{formulation3} across all possible values of $\lambda$ under the assumption that all points are projected, preserving the same unit direction and $v_{\hat{\jmath}}$=1 for the preserved direction $\hat{\jmath}$.  
\end{prop}   

Once a component has been derived, either by specifying $\lambda$ in Algorithm \ref{alg1} or choosing from among solutions produced by Algorithms 2 and 3, successive components can be computed to estimate higher-dimensional fitted subspaces.  Algorithm \ref{alg1} (or Algorithms 2 and 3) can be applied to the data projected in the null space of the subspace defined by the components derived thus far. 
 
 \section{Experiments with Synthetic Data}
In this section, we shall first analyze a toy sample to illustrate the complete solution path in terms of breakpoints and coordinate preservation, that is, how breakpoints affect solutions by changing preserved coordinates $\hat{\jmath}$. Next, we conduct simulation studies to evaluate the performance of Algorithm \ref{alg1} against methods for sparse and robust PCA.  Error is measured by the discordance between the vector $v$ defining the ``true'' line and the vector $v^*$ derived by Algorithm \ref{alg1} or a competing method.  Sparsity is measured by the $\ell_0$ norm of the solution vector $v$ which is the number of non-zero coordinates.

\subsection{A Toy Example}
\label{sec:toy}
Let us first consider five points $(4,-2,3,-6)^T$, $(-3,4,2,-1)^T$, $(2,3,-3,-2)^T$, $(-3,4,2,3)^T$, $(5,3,2,-1)^T$. Algorithm 2 generates the following breakpoints for $\lambda$: $\{1, 3\}$ for $\hat{\jmath}=1$, $\{4, 6\}$ for $\hat{\jmath}=2$, $\{0, 2\}$ for $\hat{\jmath}=3$ and $\{3, 5, 11\}$ for $\hat{\jmath}=4$. The collection of breakpoints for $\lambda$ is $\{0,1,2,3,4,5,6,11\}$. We now illustrate that the optimal solution (under the assumption that all points preserve the same coordinate $\hat{\jmath}$ and $v_{\hat{\jmath}} = 1$) might change due to the existence of additional breakpoints between successive breakpoints generated from Algorithm 2. Algorithm 3 iterates by preserving $j=1,2,3$ to find the lowest objective value over the interval $(3.5,4]$ for $\lambda$, giving rise to an additional breakpoint $3.5$. The value of the objective function comprises the error term ($\sum_{i=1}^n\|x_{i}-vx_{i\hat{\jmath}}\|_1$) and the penalty term ($\|v\|_1$), both fixed over each interval for each coordinate $j$. Algorithm 2 finds all possible breakpoints without filtering comparable larger objective function values, which is assessed in Algorithm 3. In other words, Algorithm 3 further narrows the breakpoint intervals of Algorithm 2 by evaluating $m$ objective function values. The complete solution path is summarized in Table \ref{tab:Of}. 
\begin{table}[!t]  
\centering
  \caption{Solution Path for Toy Example. The Best-Fit Line Is Fixed within Each of the Four Intervals for $\lambda$.}  
  \begin{tabular}{r|r|r} 
     $\lambda$&$z^*(\lambda)$& $v^*(\lambda)$\\
    \hline 
   (0.0, 3.0)   & (34.5, 42.0) & (-0.7,0.3,-0.5,1.0)\\ 
   (3.0, 3.5)   & (42.0, 42.9) & ($-\frac{2}{3}$,$\frac{1}{3}$,0.0,1.0)\\  
   (3.5, 11) & (42.9, 52.0) &  (1.0,0.0,0.0,-0.2)  \\ 
        (11, $\infty$) & (52.0, $\infty$) & (1.0,0.0,0.0,0.0) \\
\end{tabular} 
\label{tab:Of}
\end{table}

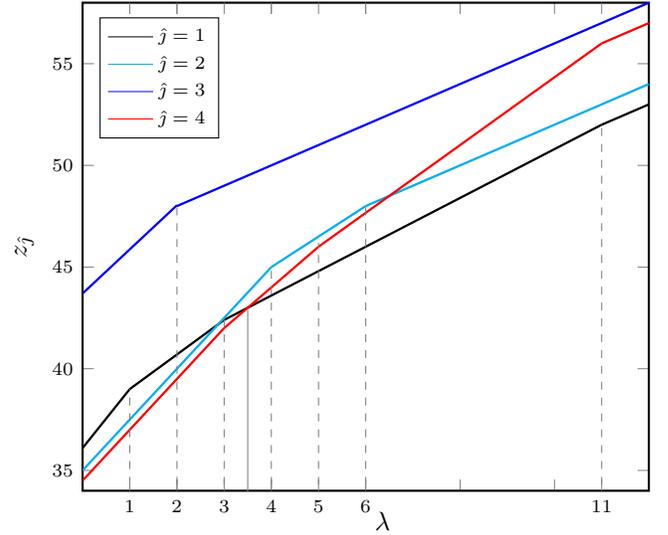
\begin{figure} [!t] 
\centering 
\begin{tikzpicture}  
\begin{axis}[%
    width=0.85\columnwidth,
    scale only axis,
    compat=newest,
    ylabel style = {align=left},
    axis line style=thick,
    inner axis line style={=>},
    xlabel={$\lambda$},  
    ylabel={$z_{\hat{\jmath}}$}, 
      x label style={anchor=west},
        y label style={anchor=south},
    ymin=34,ymax=58,
    legend pos=north west,
    xmin=0,xmax=12,
    xticklabels={},
    extra x ticks = {1,2,3,4,5,6,11},
    ticklabel style = {font=\scriptsize},
    legend style={font=\scriptsize},
    legend entries={$\hat{\jmath}=1$,
    $\hat{\jmath}=2$,
    $\hat{\jmath}=3$,
    $\hat{\jmath}=4$
    }
]
\addplot[line width=0mm ]{0};
\addplot[line width=0mm,cyan]{0}; 
\addplot[line width=0mm,blue]{0};  
\addplot[line width=0mm,color=red]{0};

\addplot[line width=0.3mm,domain=0:1 ]{36.1+2.9*x};
\addplot[line width=0.3mm,domain=1:3 ]{37.3+1.7*x}; 
\addplot[line width=0.3mm,domain=3:11 ]{38.8+1.2*x};  
\addplot[line width=0.3mm,domain=11:12 ]{41+1*x};

\addplot[line width=0.3mm,domain=0:4, cyan]{35+2.5*x};
\addplot[line width=0.3mm,domain=4:6, cyan,]{39+1.5*x};  
\addplot[line width=0.3mm,domain=6:12, cyan,]{42+1*x};  

\addplot[line width=0.3mm,domain=0:2, blue]{43.7+2.17*x};
\addplot[line width=0.3mm,domain=2:12, blue]{46+x}; 

\addplot[line width=0.3mm,domain=0:3,red,]{34.5+2.5*x};
\addplot[line width=0.3mm,domain=3:5,red,]{36+2*x};    
\addplot[line width=0.3mm,domain=5:11,red,]{37.66666+1.66666*x};
\addplot[line width=0.3mm,domain=11:12,red]{45+x};     

\draw[dashed,gray] (1,0) -- (1,39);
\draw[dashed,gray] (2,0) -- (2,48);
\draw[dashed,gray] (3,0) -- (3,42.4);
\draw[dashed,gray] (4,0) -- (4,45);
\draw[dashed,gray] (5,0) -- (5,46.2);
\draw[dashed,gray] (6,0) -- (6,48);
\draw[dashed,gray] (11,0) -- (11,52);

\draw[gray] (3.5,0) -- (3.5,42.9);
\end{axis}
\end{tikzpicture}
\caption{Schematic illustration of breakpoints. Each color depicts the objective function value when preserving a coordinate $\hat{\jmath}$, $z_{\hat{\jmath}}$, as a function of the penalty parameter $\lambda$.} 
\label{fig:ex-b}
\end{figure} 

The objective function $z_{\hat{\jmath}}$ is a linear function with respect to $\lambda$ over a certain interval for each preserved $\hat{\jmath}$. The intercept is $\sum_{i=1}^n \|x_{i}-vx_{i\hat{\jmath}}\|_1$, and $\|v\|_1$ is the slope. Figure \ref{fig:ex-b} shows that the smallest $z_{\hat{\jmath}}$ can be achieved by preserving direction $\hat{\jmath}=4$ for $\lambda$ in the interval $[0,3.5]$, depicted by the red line segment, and direction $\hat{\jmath}=1$ for $\lambda$ in the interval $[3.5,\infty)$, depicted by the black line segment.  Algorithm 2 finds all 7 breakpoints $\lambda$s at which the slopes of the lines of the same color change. Algorithm 3 finds an additional breakpoint in the interval (3,4], depicted by the gray vertical line, at which value two line segments intersect.  
At this value, the preserved direction changes from $\hat{\jmath}=4$ to $\hat{\jmath}=1$ as $\lambda$ increases. 

This result is consistent with that of Table \ref{tab:Of}, where the first two solutions preserve $\hat{\jmath}=4$ and the last two solutions preserve $\hat{\jmath}=1$. The solutions for each choice of $\lambda$ are summarized in Table \ref{tab:Of}. 

Positive regularization terms $\lambda$ are used to control the level of sparsity in the solution. And there shall be a maximum $\lambda$ beyond which all elements are 0 except for $v_{\hat{\jmath}}$ which is fixed at 1.  In this case, the maximum value value for $\lambda$ is 11.

\subsection{Comparison to Competing Methods}
In this experimental section, we compare the discordance of our Algrothm 1 to two $\ell_1$-norm PCA algorithms, namely, Sparse Robust Principal Components using the Grid search algorithm implemented in the R package pcaPP \cite{croux2013robust}, and Sparse robust PCA algorithm based on the ROBPCA \cite{hubert2005robpca} algorithm implemented in the R package rosPCA \cite{hubert2016sparse}, with respect to the $\ell_0$-norm, utilizing synthetic data. We generated a total of 30 datasets, with each set containing 100 observations in 100 dimensions. Five datasets for each of the outlier counts ranging from 1 to 6. Each set of five datasets with the same outlier count was generated using different random seeds to ensure variability. For each replication, each coordinate of the true $v$ is sampled from a Uniform(-10, 10) distribution and after all coordinates have been sampled, $v$ is normalized so that $\|v\|_2=1$. For non-outlier observations, each ``true'' $\alpha$ is sampled from a Uniform(-100,100) distribution to locate the projection on the line. Noise following a Laplace(0,10) distribution is added to locate the points off of the line. Outlier observations are created by sampling the first five coordinates from a Uniform(50,100) distribution and noise sampled form a Laplace(0,1) distribution is added.

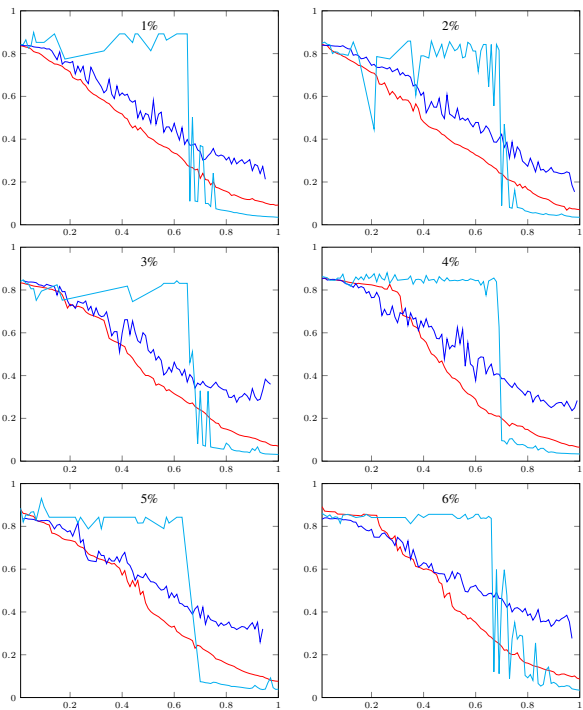
\begin{figure}[h!]  
\centering 
\scalebox{0.5}{
\begin{tikzpicture}
\pgfplotsset{ 
    xmin=0.01, xmax=1
}
\begin{axis}[ 
    ymin=0, ymax=1,  
    ticklabel style = {font=\scriptsize},
    title style={anchor=north,yshift=-10},
    title = 1\%,
] 
\addplot[help lines,mark size=2pt,color=red]
    table [x=l0norm, y=l1pca0.01,col sep=comma]{csv/rspcaadj.csv};   
\addplot[help lines,color=blue,mark size=2pt]
    table [x=l0norm, y=pppca0.01,col sep=comma]{csv/rspcaadj.csv};
\addplot [help lines,color=cyan,mark size=2pt]
    table [x=l0norm, y=rospca0.01,col sep=comma]{csv/rspcaadj.csv};     
    \coordinate (s1) at (axis cs:0.95,0);
    \coordinate (s2) at (axis cs:0.8,0.4);
\end{axis} 
\end{tikzpicture}}
\scalebox{0.5}{
\begin{tikzpicture}
\pgfplotsset{ 
    xmin=0.01 , xmax=1
}
\begin{axis}[ 
    ymin=0, ymax=1,  
    ticklabel style = {font=\scriptsize},
    title style={anchor=north,yshift=-10},
    title = 2\%,
] 
\addplot[help lines,very thin,color=red]
    table [x=l0norm, y=l1pca0.02,col sep=comma] {csv/rspcaadj.csv};   
\addplot[help lines,color=blue,very thin]
    table [x=l0norm, y=pppca0.02,col sep=comma] {csv/rspcaadj.csv};
\addplot[help lines,color=cyan,very thin]
    table [x=l0norm, y=rospca0.02,col sep=comma] {csv/rspcaadj.csv};    
        \coordinate (s1) at (axis cs:0.95,0);
    \coordinate (s2) at (axis cs:0.8,0.4);
\end{axis}  
\end{tikzpicture}}

\scalebox{0.5}{
\begin{tikzpicture}
\pgfplotsset{ 
    xmin=0.01 , xmax=1
}
\begin{axis}[ 
    ymin=0, ymax=1,  
    ticklabel style = {font=\scriptsize},
    title style={anchor=north,yshift=-10},
    title = 3\%,
] 
\addplot[help lines,very thin,color=red]
    table [x=l0norm, y=l1pca0.03,col sep=comma] {csv/rspcaadj.csv};   
\addplot[help lines,color=blue,very thin]
    table [x=l0norm, y=pppca0.03,col sep=comma] {csv/rspcaadj.csv};
\addplot [help lines,color=cyan,very thin]
    table [x=l0norm, y=rospca0.03,col sep=comma] {csv/rspcaadj.csv};     
     \coordinate (s1) at (axis cs:0.95,0);
    \coordinate (s2) at (axis cs:0.8,0.4);
\end{axis}  
\end{tikzpicture}} 
\scalebox{0.5}{
\begin{tikzpicture}
\pgfplotsset{ 
    xmin=0.01, xmax=1
}
\begin{axis}[ 
    ymin=0, ymax=1,  
    ticklabel style = {font=\scriptsize},
    title style={anchor=north,yshift=-10},
    title = 4\%,
] 
\addplot[help lines, color=red,very thin]
    table [x=l0norm, y=l1pca0.04,col sep=comma] {csv/rspcaadj.csv};   
\addplot[help lines, color=blue,very thin]
    table [x=l0norm, y=pppca0.04,col sep=comma] {csv/rspcaadj.csv};
\addplot [help lines, color=cyan,very thin]
    table [x=l0norm, y=rospca0.04,col sep=comma] {csv/rspcaadj.csv};    
  \coordinate (s1) at (axis cs:0.95,0);
    \coordinate (s2) at (axis cs:0.8,0.4);
\end{axis}  
\end{tikzpicture}}

\scalebox{0.5}{
\begin{tikzpicture}
\pgfplotsset{ 
    xmin=0.01, xmax=1
}
\begin{axis}[ 
    ymin=0, ymax=1,  
    ticklabel style = {font=\scriptsize},
    title style={anchor=north,yshift=-10},
    title = 5\%,
] 
\addplot[help lines, very thin,color=red]
    table [x=l0norm, y=l1pca0.05,col sep=comma] {csv/rspcaadj.csv};   
\addplot[help lines,color=blue,very thin]
    table [x=l0norm, y=pppca0.05,col sep=comma] {csv/rspcaadj.csv};
\addplot [help lines,color=cyan,very thin]
    table [x=l0norm, y=rospca0.05,col sep=comma] {csv/rspcaadj.csv};     
  \coordinate (s1) at (axis cs:0.95,0);
    \coordinate (s2) at (axis cs:0.8,0.5);
\end{axis}  
\end{tikzpicture}}
\scalebox{0.5}{
\begin{tikzpicture}
\pgfplotsset{ 
    xmin=0.01, xmax=1
}
\begin{axis}[ 
    ymin=0, ymax=1,  
    ticklabel style = {font=\scriptsize},
    title style={anchor=north,yshift=-10},
    title = 6\%,
] 
\addplot[help lines, very thin,color=red]
    table [x=l0norm, y=l1pca0.06,col sep=comma] {csv/rspcaadj.csv};   
\addplot[help lines,color=blue,very thin]
    table [x=l0norm, y=pppca0.06,col sep=comma] {csv/rspcaadj.csv};
\addplot [help lines,color=cyan,very thin]
    table [x=l0norm, y=rospca0.06,col sep=comma] {csv/rspcaadj.csv};   
  \coordinate (s1) at (axis cs:0.95,0);
    \coordinate (s2) at (axis cs:0.8,0.5);
\end{axis}  
\end{tikzpicture}}
\caption{The figure displays the discordance (y-axis) versus the $\ell_0$-norm (x-axis) for Algorithm 1 (red), pcaPP (blue), and rosPCA (cyan) on simulated data with increasing levels of contamination. The plots compare the estimated loadings matrix to the true loadings matrix for data with outlier levels ranging from 1\% to 6\%. }
\label{fig:rscomp}
\end{figure}

Figure \ref{fig:rscomp} presents six plots comparing the performance of three algorithms in terms of mean discordance on the y-axis against the mean $\ell_0$-norm on the x-axis  (averaged across five datasets), for varying outlier counts. All algorithms demonstrate a decrease in discordance with an increase in $\ell_0$-norm of the solution vector, suggesting a higher challenge in handling sparser data. Results show pcaPP and rosPCA discordance are lower than Algorithm 1 in some sparser instance, Algorithm 1 exhibits a steady trend throughout the range of $\ell_0$ norm, evidenced by consistently maintaining a lower discordance than rosPCA and pcaPP. As the outlier level increases, pcaPP produces more high-sparsity models with the lowest discordance among the three methods. Notably, rosPCA, while not as effective in terms of discordance in sparser scenarios (when $\ell_0$-norm is less than 0.7), outperforms Algorithm 1 and pcaPP in less sparse situations (when $\ell_0$-norm is greater than 0.7). Overall, we observe that Algorithm 1 often produces models with the lowest discordance and exhibits the smoothest trade-off in discordance for changes in sparsity. The consistency suggests that Algorithm 1 is more robust in response to variations in levels of outliers. 

\subsection{Effect of Varying $m$ and $n$ on Error and Sparsity}
In Figure \ref{fig:ll1}(a), we illustrate the impact of varying $\lambda$ under two scenarios without the presence of outliers. First, we investigate the variation of the number of columns (m) across three settings: $m$ = 1000, 2000, and 3000, while keeping the number of rows $n$ fixed at 3000. Secondly, we explore the effect of changing the number of rows $n$ across 1000, 2000, and 3000, while fixing a constant columns $m$ at 1000. 

To ensure diversity in our analysis, we generated each set of five datasets of the same size using different random seeds. In each replication, each coordinate of the true $v$ is sampled from a uniform(-1, 1) distribution and after all coordinates have been sampled, $v$ is normalized so that $\|v\|_2=1$. Each ``true'' $\alpha$ is sampled from a uniform distribution (-100,100) to locate the projection on the line. As shown in Figure \ref{fig:ll1}(a), the fitted line exhibits a higher rate of change in its response to $\lambda$ in terms of discordance and the $\ell_0$ norm (averaged over five datasets), as the dimensionality $m$ increases with the number of observations $n$ kept at 3000. In contrast, with a fixed dimensionality $m$, the fitted line shows a higher rate of change in response to $\lambda$ regarding discordance and the $\ell_0$ norm as the number of observations $n$ decreases, as shown in Figure \ref{fig:ll1}(b). It also shows that a $\lambda$ less than the intersection point (between the discordance curve and the $\ell_0$ norm curve) will lead to a solution with discordance below 0.4 and sparsity below 40\%. Figure \ref{fig:ll1}(b) shows that solutions with similar properties can be obtained by varying $n$.  

\begin{figure} [!t]  
\centering 
\begin{tikzpicture}[scale=1]
\pgfplotsset{ 
    xmin=0, xmax=8500
}
\begin{axis}[
    xlabel={$\lambda$},
    ylabel={Discordance/$\ell_0$},  
    legend style={draw=none},
    ymin=0, ymax=1, 
    legend style={at={(0.4,0.4)},anchor=north east},
    ymajorgrids=true,
    grid style=dashed,
    ticklabel style = {font=\scriptsize},
] 
\addplot[color=blue,mark=o,]
    table [x=lambda, y=1000d,col sep=comma] {csv/lbehavior_fixedrowsat3000.csv};
    \legend{3000x1000}
\addplot[color=red,mark=square,]
    table [x=lambda, y=2000d,col sep=comma] {csv/lbehavior_fixedrowsat3000.csv};
    \addlegendentry{3000x2000}
\addplot[color=green,mark=triangle,]
    table [x=lambda, y=3000d,col sep=comma] {csv/lbehavior_fixedrowsat3000.csv};
    \addlegendentry{3000x3000}
    
\addplot[color=blue,mark=o,dashed,mark options={solid}]
    table [x=lambda, y=1000l0,col sep=comma] {csv/lbehavior_fixedrowsat3000.csv};  
\addplot[color=red, dashed,mark=square,mark options={solid}]
    table [x=lambda, y=2000l0,col sep=comma] {csv/lbehavior_fixedrowsat3000.csv};  
\addplot[color=green,dashed,mark=triangle,mark options={solid}]
    table [x=lambda, y=3000l0,col sep=comma] {csv/lbehavior_fixedrowsat3000.csv};  
\end{axis}  
\end{tikzpicture}

(a)

\begin{tikzpicture}[scale=1]
\pgfplotsset{ 
    xmin=0, xmax=5500
}
\begin{axis}[
    xlabel={$\lambda$}, 
    legend style={draw=none},
    ymin=0, ymax=1, 
    legend style={at={(0.4,0.4)},anchor=north east},
    ymajorgrids=true,
    grid style=dashed,
] 
\addplot[color=blue,mark=o,]
    table [x=lambda, y=1000d,col sep=comma] {csv/lbehavior_fixedcolsat1000.csv};
    \legend{1000x1000}
\addplot[color=red,mark=square,]
    table [x=lambda, y=2000d,col sep=comma] {csv/lbehavior_fixedcolsat1000.csv};
    \addlegendentry{2000x1000}
\addplot[color=green,mark=triangle,]
    table [x=lambda, y=3000d,col sep=comma] {csv/lbehavior_fixedcolsat1000.csv};
    \addlegendentry{3000x1000}
\end{axis}  
\begin{axis}[  
    ylabel={Discordance/$\ell_0$}, 
    ymin=0.0003, ymax=1,  
    yticklabel pos=left,    
]
\addplot[color=blue,mark=o,dashed,mark options={solid}]
    table [x=lambda, y=1000l0,col sep=comma] {csv/lbehavior_fixedcolsat1000.csv};  
\addplot[color=red, dashed,mark=square,mark options={solid}]
    table [x=lambda, y=2000l0,col sep=comma] {csv/lbehavior_fixedcolsat1000.csv};  
\addplot[color=green,dashed,mark=triangle,mark options={solid}]
    table [x=lambda, y=3000l0,col sep=comma] {csv/lbehavior_fixedcolsat1000.csv};  
\end{axis} 
\end{tikzpicture}

(b)

\caption{Effect on discordance (solid lines) and sparsity (dashed lines) when $\lambda$ is varied for (a) datasets with different $m$ and (b) datasets with different $n$.  Sparsity is measured as a percentage of $m$.}
\label{fig:ll1}
\end{figure}

\section{Application to Human Microbiome Project Data}
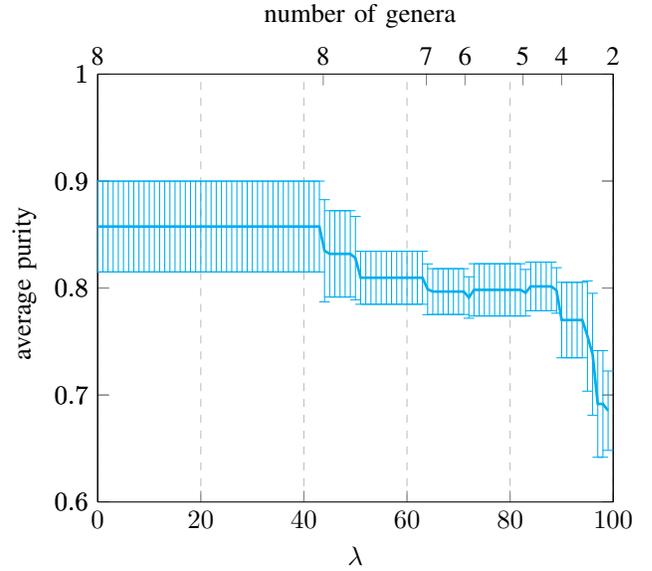
\begin{figure} [!t]  
\pgfplotsset{set layers}
\begin{tikzpicture}[scale=1]
\pgfplotsset{ 
    ymin=0.6, ymax=1
}

\begin{axis}[ 
    xlabel={$\lambda$},
    ylabel={\ average purity},  
    axis x line*=bottom, 
    xmin=0, xmax=100,   
    xmajorgrids=true,
    grid style=dashed, 
] 
\addplot[line width=1pt,color=cyan,mark=none]
plot[error bars/.cd,y dir=both,y explicit]
    table [x=lambda, y=avg_purity,y error=sd_purity,col sep=comma]{csv/hmp.csv}; 
\end{axis}  

\begin{axis}[  
    axis x line*=top,
    xmin=0, xmax=8,  
    xticklabel pos=top,   
    xlabel={\ number of genera}, 
    xtick={0,3.5,5.1,5.7,6.6,7.2,8},
    xticklabels={8,8,7,6,5,4,2}
] 
\end{axis} 
\end{tikzpicture} 
\caption{The number of taxa used for clustering is on the left y-axis for the blue line and average purity is on the right y-axis for the red line.}
\label{fig:purity}
\end{figure}

This section demonstrates the efficacy of Algorithm 1 in selecting pertinent features, thereby achieving commendable clustering results. This experiment was carried out using data from the Human Microbiome Project \cite{peterson2009nih}. The dataset comprises relative abundances of 2,355 tissue samples collected from four distinct body sites: the skin, gut, oral cavity, and vagina. To ensure a balanced representation, we randomly selected 234 samples from each site, totaling 968 samples, and performed 10 shufflings. A set of 320 genera associated with each tissue sample was entered into Algorithm 1, using a $\lambda=0$ to $\lambda=99$. This process produced sparse components (1st component for each $\lambda$), corresponding to the active subset of genera. Hierarchical clustering was then conducted using an average merging strategy with the Jaccard distance as a metric. The performance of the clustering was evaluated by purity, defined as $\frac{1}{968}\sum_{i=1}^{r}\max_{j=1}^k{c_{ij}}$, for the body sites $k=4$ and the groups $r=4$. Here, $c_{ij}$ represents the size of the group within each cluster that shares the most samples with a given body site. As illustrated in Figure \ref{fig:purity}, the purity consistently maintains a high level (86\%) while $\lambda$ varies from 0 to 99, and the number of genera used for clustering decreases from 8 to 2. With the incremental increase in $\lambda$, a discernible pattern emerges in the removal of features from the clustering model. Bacteroides is the first genus to be excluded, followed by Rothia, Prevotella, Veillonella, Streptococcus, Haemophilus, and Corynebacterium. The process ends with the two features, Lactobacillus and Corynebacterium, remaining in the model. Excluding Bacteroides results in an average purity drop from 86\% to 80\%. Comparatively, only using 8 core genera (Table \ref{tab:hmp}) achieving an average 86\% hierarchical clustering purity across 10 shufflings. The eight genera were included in the 51 core genera listed in \cite{tan2021machine} Table S2. Figure \ref{fig:purity} shows the purity behavior in relation to $\lambda$. In particular, Lactobacillus and Corynebacterium emerges particularly significant discriminators. It is noteworthy that four genera are still capable of achieving an average purity of 80\%. The purity of hierarchical clustering reaches 60\% when exclusively using them.
\begin{table}[ht]
\centering
\caption{Core genera identified by Algorithm 1 with $\lambda$=0.}
\label{tab:hmp}
\begin{tabular}{l|l|l|l}
\toprule
Streptococcus  &   Corynebacterium            & Haemophilus    & Veillonella \\
Rothia &  Bacteroides         &Prevotella &Lactobacillus\\
\bottomrule
\end{tabular}
\end{table}

\section{Algorithms 1 and 2 on NVIDIA Graphical Processing Units}

\subsection{Introduction}
Algorithms \ref{alg1} and \ref{alg2} can be implemented in a parallel framework CUDA by sorting the $m$ lists independently. 
The CUDA API enables C/C++ to execute parallel algorithms within thread groups on GPUs. It combines sequential and parallel abstractions, i.e., Host on CPU and Device on GPU. Users launch an application in the Host, connecting to Device through a kernel interface. Devices are responsible for performing computations and partitioning the cache.   As depicted schematically in Figure \ref{fig:cuda2}, two blocks, each consisting of four threads, execute parallel computations on vector elements cached within the GPU. Memory locations of elements can be accessed using four built-in variables \texttt{threadIdx.x}, \texttt{blockIdx.x}, \texttt{blockDim.x}, and \texttt{gridDim.x}. Specifically, \texttt{threadIdx.x} represents the index of each element within a block, \texttt{blockIdx.x} denotes the index of each block, \texttt{gridDim.x} is the total number of blocks, and \texttt{blockDim.x} indicates the number of threads per block. 
 \begin{figure} [!htb]  
 \centering
 \small
\begin{tikzpicture} 
[th/.style={rectangle,draw},
px/.style={rectangle,rounded corners, draw = black!70 },
line/.style={draw, -latex'}]  
  \node[px] at (-2.5,1.5)(x1){$x_{0}$};
  \node[left=1 of x1,text=black](v1){Host};
    \node[px,right =0.1 of x1](x2){$x_{1}$};
    \node[px,right =0.1 of x2](x3){$x_{2}$};
    \node[px,right =0.1 of x3](x4){$x_{3}$};
    \node[px,right =0.1 of x4](x5){$x_{4}$};
    \node[px,right =0.1 of x5](x6){$x_{5}$};
    \node[px,right =0.1 of x6](x7){$x_{6}$};
    \node[px,right =0.1 of x7](x8){$x_{7}$};
    \node[below=0.2 of v1,text=black](k1){Kernel}; 
    \node[right= 2.6 of k1](k2){$\lll2,4\ggg$};
     \node[below=0.4 of k1,text=black](g1){Device}; 
    \node[th, right = 2 of g1] (t1) {thread 0};
    \node [th,below =0.1 of t1] (t2) {thread 1}; 
    \node [th,below =0.1 of t2] (t3) {thread 2};
    \node [th,below =0.1 of t3] (t4) {thread 3}; 
    \node [th,right =0.6 of t1] (t5) {thread 0}; 
    \node [th,below =0.1 of t5] (t6) {thread 1};
    \node [th,below =0.1 of t6] (t7) {thread 2};
    \node [th,below =0.1 of t7] (t8) {thread 3};
    \node[below=0.3 of t4,text=black]{$0$};
    \node[below=0.3 of t8,text=black]{$1$}; 
\draw[rounded corners,very thick] (-1.8,0.4)  rectangle  (-0.2,-2);
\draw[rounded corners,very thick] (0.1,0.4)   rectangle   (1.7,-2);
\node[below=1 of t4](f1){$x_0:threadIdx.x+blockIdx.x\cdot blockDim.x=0+0\cdot2$};  
\end{tikzpicture} 
\caption{\textbf{One dimensional decomposition using blocks and threads. The formula will map each thread to an element in the vector.}}   \label{fig:cuda2}
\end{figure}

Our CUDA experiments run on an NVIDIA GeForce RTX 3060 mobile GPU, equipped with 3840 cores and 6 gigabytes of graphics memory.  CPU implementations are executed on an Intel 8-core i9 processor with 40 gigabytes of memory. 

\subsection{Algorithm \ref{alg1} Performance Evaluation}
We execute the CUDA implementation with ten replications for each size and calculate the average runtime.
\begin{table} 
\caption{Average and standard deviation times in seconds for 10 replications of each dataset with a varying number of columns and a fixed number of rows at 1000, 2000, and 5000.}
\label{tab:la1}  
\centering
\csvreader[
tabular = r|rrr,
table head = & \multicolumn{3}{c}{Number of Rows}\\ & 1000 & 2000 & 5000\\
\hline,
late after line = \\]%
{csv/sparsel1.csv}{}{%
\csvcoli&\csvcolv~$_{\csvcolxxiii}$&\csvcolvi~$_{\csvcolxxiv}$ & \csvcolvii~$_{\csvcolxxv}$
}  
\end{table}
Table \ref{tab:la1} presents the average elapsed time for computing Algorithm \ref{alg1} across 10 replications for various matrix sizes.  It is noteworthy that completing the analysis of a $5000\times1000$ matrix on the GPU in just 26 seconds. As expected, the running time of Algorithm \ref{alg1} is directly proportional to both the $m^2$ and $n$. Consequently, larger matrices with more columns necessitate increased computation time compared to matrices with fewer columns. This relationship is further illustrated in Table \ref{tab:la2}, where a matrix with 5000 columns required 127 seconds for analysis.

Table \ref{tab:speedup1} presents an overview of the speedup for 121 different input sizes, demonstrating up to a 16.57x improvement over the CPU implementation. This indicates a clear, increasing speedup trend as the size increases.

\begin{table} 
\caption{Average and standard deviation times in seconds over 10 replications for each dataset with varying the number of rows and a fixed number of columns at 1000, 2000, and 5000. }
\label{tab:la2}   
\centering
\csvreader[
tabular = r|rrr,
table head = & \multicolumn{3}{c}{Number of Columns}\\ & 1000 & 2000 & 5000\\
\hline,
late after line = \\]%
{csv/sparsel1.csv}{}{%
\csvcoli&\csvcolii~$_{\csvcolxx}$&\csvcoliii~$_{\csvcolxxi}$ & \csvcoliv~$_{\csvcolxxii}$
} 
\end{table}

\begin{table}[!htb] 
\caption{Speedup Results for a Matrix of Dimension Row Index $\times$ Column Header. A Value Greater than 1 Demonstrates the Increased Efficiency of the Parallel Implementation of Algorithm \ref{alg1}.}
\csvreader[
tabular = r|rrrrrrrrrrr,
table head = & \multicolumn{10}{c}{Number of Columns} \\ &100&200&300&400&500&600&700&800&900&1000&2000 \label{tab:speedup1}\\
\hline,
late after line = \\]%
{csv/speedup.csv}{}{%
\csvcoli&\csvcolii&\csvcoliii&\csvcoliv&\csvcolv&\csvcolvi&\csvcolvii&\csvcolviii&\csvcolix&\csvcolx&\csvcolxi&\csvcolxii
}
\end{table}

\subsection{Breakpoints for Algorithm 2}
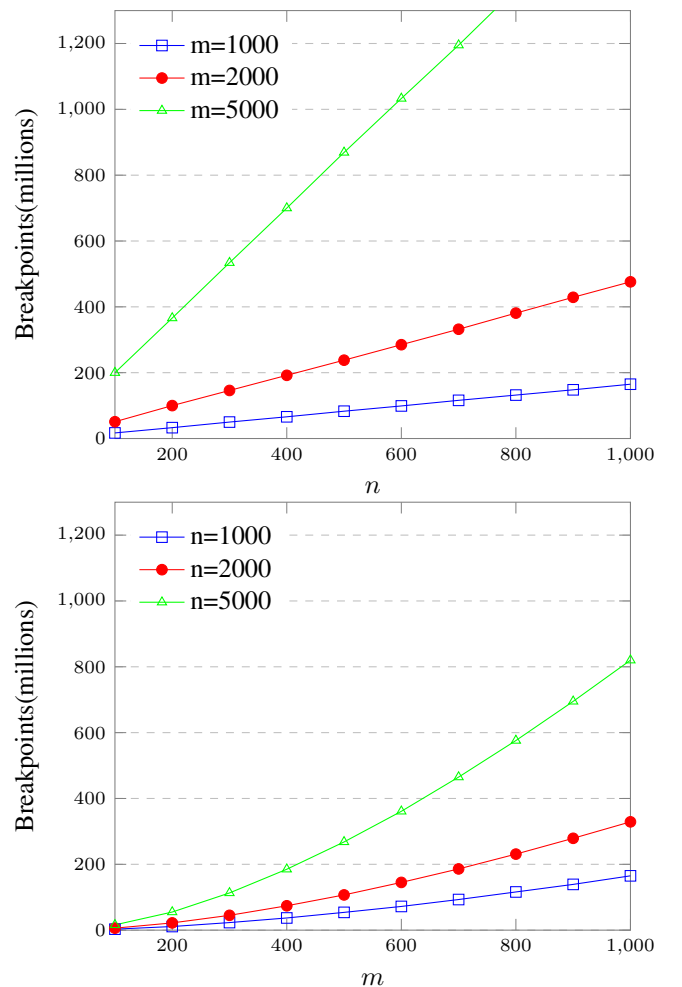
\begin{figure} [!ht]
\centering 
\begin{tikzpicture}[scale=1]
\begin{axis}[
    xlabel={$n$},
    ylabel={Breakpoints(millions)},
    xmin=100, xmax=1000,
    ymin=0, ymax=1300, 
    legend pos= north west,
    ymajorgrids=true,
    grid style=dashed,
    axis line style={gray},
    legend style={draw=none},
    ticklabel style = {font=\scriptsize},
]
\addplot[
    color=blue,
    mark=square,
    ]
    table [x=x, y=lm1000,col sep=comma] {csv/sparsel1.csv};
    \legend{m=1000}
\addplot[
    color=red,
    mark=*,
    ]
    table [x=x, y=lm2000,col sep=comma] {csv/sparsel1.csv};
    \addlegendentry{m=2000}
\addplot[
    color=green,
    mark=triangle,
    ]
    table [x=x, y=lm5000,col sep=comma] {csv/sparsel1.csv};
    \addlegendentry{m=5000}
\end{axis}
\end{tikzpicture} 
\begin{tikzpicture}[scale=1]
\begin{axis}[ 
    xlabel={$m$}, 
    ylabel={Breakpoints(millions)},
    xmin=100, xmax=1000,
    ymin=0, ymax=1300, 
    legend pos= north west,
    ymajorgrids=true,
    grid style=dashed,
    axis line style={gray},
    legend style={draw=none},
    ticklabel style = {font=\scriptsize},
]
\addplot[
    color=blue,
    mark=square,
    ]
    table [x=x, y=ln1000,col sep=comma] {csv/sparsel1.csv};
    \legend{n=1000}
\addplot[
    color=red,
    mark=*,
    ]
    table [x=x, y=ln2000,col sep=comma] {csv/sparsel1.csv};
    \addlegendentry{n=2000}
\addplot[
    color=green,
    mark=triangle,
    ]
    table [x=x, y=ln5000,col sep=comma] {csv/sparsel1.csv};
    \addlegendentry{n=5000}
\end{axis}
\end{tikzpicture}
\caption{Number of breakpoints for Algorithm \ref{alg2}. The x-axis begins at 100 for both plots.}
\label{fig:bp1}
\end{figure}

In this section, we evaluate the space requirements for the number of breakpoints generated by Algorithm 2, as shown in Figure \ref{fig:bp1}. 
We fixed the column counts at 1000, 2000, and 5000, while varying the number of rows (as shown in the left panel). An upward trend is observed, indicating that an increase in the number of rows leads to a corresponding rise in breakpoints. Notably, the line corresponding to $m=5000$ is steeper than those for the other two column counts, suggesting a more significant impact on breakpoints with a higher column count. An upward trend persists when fixing the row counts at 1000, 2000, and 5000 and varying the number of columns (as shown in the right panel). The line for $n=5000$ is markedly steeper, indicating that a more significant number of columns significantly influences the number of breakpoints. In each plot, the number of breakpoints grows more rapidly as the fixed values of $m$ or $n$ increase. Algorithm 2 demonstrates a greater sensitivity to variations in $m$ (the number of columns) compared to $n$ (the number of rows), evidenced by sharper inclines in the first plot relative to the second when holding other value constant. Finally, the plot suggests a linear relationship between the number of breakpoints and the number of columns ($n$) and a higher-order polynomial relationship with the number of rows ($m$).

\begin{table} [ht!]
\caption{Average and standard deviation of breakpoints in millions over 10 replications with varying numbers of columns and a fixed number of rows at 1000, 2000, and 5000.}
\centering 
\csvreader[
tabular = r|rrr,
table head =  & \multicolumn{3}{c}{Number of Rows} \\ & 1000 & 2000 & 5000\label{tab:breakpoints2}\\
\hline,
late after line = \\]%
{csv/sparsel1.csv}{}{%
\csvcoli&\csvcolxvii~$_{\csvcolxxxv}$&\csvcolxviii~$_{\csvcolxxxvi}$ & \csvcolxix~$_{\csvcolxxxvii}$
} 
\end{table}

\begin{table}[ht!] 
\caption{Average and standard deviation of breakpoints in millions over 10 replications with varying the number of rows and a fixed number of columns at 1000, 2000, and 5000.}
\centering 
\csvreader[
tabular = r|rrr,
table head =  & \multicolumn{3}{c}{Number of Columns} \\ & 1000 & 2000 & 5000\label{tab:breakpoints3}\\
\hline,
late after line = \\]%
{csv/sparsel1.csv}{}{%
\csvcoli&\csvcolxiv~$_{\csvcolxxxii}$&\csvcolxv~$_{\csvcolxxxiii}$ & \csvcolxvi~$_{\csvcolxxxiv}$
}
\end{table}

\section{Conclusion}
We introduce an innovative and effective approach to fit a one-dimensional subspace that addresses the challenges of outliers, scalability, and sparsity. The proposed approach formulates the fitting procedure as a mathematical optimization problem, employing the $\ell_1$-norm to determine the best-fit subspace. Additionally, it imposes an $\ell_1$-norm penalty term to induce sparsity and produce sparse principal components successively. This approach is based on the principles of linear programming. It is shown that solving pertinent linear programming can be accomplished using simple ratios and sorting. Moreover, our algorithms are designed to support parallel processing, thereby enhancing their efficiency. Evidence from both simulations and analyses of actual data has shown that our robust sparse subspaces are outliers-insensitive in the data, and the sparsity patterns identified are indeed valuable. The approach was provided in conjunction with an algorithm to generate the complete solution path. Additionally, we demonstrate the relationship between representation error, sparsity, and the regularization parameter. This serves as an important exploratory tool to help identify the optimal level of sparsity. The full implementation of our algorithm is available through the R package pcaL1 \cite{pcal1} and also supports CUDA environments available in \cite{l1l1git}.

\section*{Acknowledgments}

\bibliographystyle{IEEEtran}
\bibliography{ref.bib}
  
\begin{IEEEbiography}[{\includegraphics[width=1in,height=1.25in,clip,keepaspectratio]{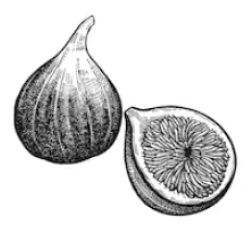}}]{Xiao Ling} received the B.S. degree in electrical engineering from University of Shanghai for Science and Technology, China, the M.S. degree in Mathematics from the University of Toledo, Ohio, and PhD degree in systems modeling and analysis from Virginia Commonwealth University, Virginia, USA.  He is currently a postdoctoral fellow at University of Maryland, Baltimore, USA. His research interests include mathematical optimization, applied statistics and bioinformatics.  
\end{IEEEbiography}

\begin{IEEEbiography}[{\includegraphics[width=1in,height=1.25in,clip,keepaspectratio]{fig1}}]{Paul Brooks} was born in Atlanta, Georgia, U.S.A., on June 13, 1977.  He earned a B.A. in mathematics and a B.A. in physics from the University of Virginia, Charlottesville, Virginia, in 1999.  He earned an M.S. in operations research in 2003 and a Ph.D. in operations research in 2005 from Georgia Institute of Technology, Atlanta, Georgia.

He was an assistant professor at Piedmont College from 2004 to 2006.  He joined Virginia Commonwealth University in 2006 as an assistant professor.  He was promoted to associate professor and then professor.  He has held appointments in the Department of Statistical Sciences and Operations Research and the Department of Supply Chain Management and Analytics.   He is currently professor and chair in the Department of Information Systems.

He is a member of Institute for Operations Research and Management Science (INFORMS) and past chair of its Data Mining Society.  He is also a member of the INFORMS Computing Society, Health Applications Society, and Optimization Society.  He is a member of the Mathematical Optimization Society (MOS).
\end{IEEEbiography}
\end{document}